\documentclass[letterpaper, 10 pt, conference]{ieeeconf}

\IEEEoverridecommandlockouts                              %

\title{\LARGE \bf

Virtual Omnidirectional Perception for Downwash Prediction within a Team of Nano Multirotors Flying in Close Proximity
}

\author{Akmaral Moldagalieva and Wolfgang Hönig%
\thanks{The authors are with TU Berlin, Germany. {\tt\footnotesize \{moldagalieva, hoenig\}@tu-berlin.de}.}%
\thanks{We thank Pablo Robles Cervantes for help with synthetic image generation and Dennis Schmidt and Khaled Wahba for help with flight experiments.}%
\thanks{Code, Data, Video: \url{https://github.com/IMRCLab/dataset-cv-rel-pos}}
\thanks{The research was funded by the Deutsche Forschungsgemeinschaft (DFG, German Research Foundation) - 448549715.
}%
}

\usepackage[bookmarks=true]{hyperref}
\usepackage{amssymb}
\usepackage{amsmath}
\usepackage{multirow}
\usepackage[detect-weight=true, detect-family=true,detect-inline-weight=math]{siunitx}

\usepackage[detect-weight=true, detect-family=true,detect-inline-weight=math]{siunitx}

\usepackage[ruled,vlined,linesnumbered]{algorithm2e}
\SetInd{0.25em}{0.5em}
\SetAlFnt{\footnotesize\sf}
\SetKwRepeat{Do}{do}{while}
\SetKw{KwStep}{step}
\SetKwInput{KwData}{Input}
\SetKwInput{KwResult}{Result}
\SetKwBlock{RepeatForever}{repeat}{}
\SetKw{Continue}{continue}
\SetKwComment{Comment}{$\triangleright$\ }{}
\SetCommentSty{itshape}

\usepackage[capitalise]{cleveref} %
\Crefname{section}{Sec.}{Secs.}
\crefname{section}{Sec.}{Secs.}
\usepackage{flushend}

\usepackage{tikz}
\usetikzlibrary{fit,calc}
\newcommand{\vp}{\mathbf{p}}    %
\newcommand{\vb}{\mathbf{b}}    %
\newcommand{\vv}{\mathbf{v}}    %
\newcommand{\vomega}{\boldsymbol{\omega}}    %
\newcommand{\vf}{\mathbf{f}}    %
\newcommand{\vtau}{\boldsymbol{\tau}}    %
\newcommand{\veta}{\boldsymbol{\eta}}    %
\newcommand{\vu}{\mathbf{u}}    %
\newcommand{\vrho}{\boldsymbol{\rho}}    %

\newcommand{\um}{\textit{m}}

\newcommand{\mT}{\mathbf{T}}    %
\newcommand{\mK}{\mathbf{K}}    %
\newcommand{\fC}{\mathcal{C}}   %
\newcommand{\fR}{\mathcal{R}}   %
\newcommand{\fW}{\mathcal{W}}   %
\newcommand{\fI}{\mathcal{I}}   %
\newcommand{\sB}{\mathcal{B}}   %

\newcommand{\vg}{\mathbf{g}}    %

\DeclareMathOperator*{\minmax}{min\,max}
\DeclareMathOperator*{\diag}{diag}

\begin{document}

\maketitle
\thispagestyle{empty}
\pagestyle{empty}

\begin{abstract}
Teams of flying robots can be used for inspection, delivery, and construction tasks, in which they might be required to fly very close to each other. In such close-proximity cases, nonlinear aerodynamic effects can cause catastrophic crashes, necessitating each robots' awareness of the surrounding. Existing approaches rely on multiple, expensive or heavy perception sensors. Such perception methods are impractical to use on nano multirotors that are constrained with respect to weight, computation, and price. Instead, we propose to use the often ignored yaw degree-of-freedom of multirotors to spin a single, cheap and lightweight monocular camera at a high angular rate for omnidirectional awareness of the neighboring robots. We provide a dataset collected with real-world physical flights as well as with 3D-rendered scenes and compare two existing learning-based methods in different settings with respect to success rate, relative position estimation, and downwash prediction accuracy. We demonstrate that our proposed spinning camera is capable of predicting the presence of aerodynamic downwash with an $\mathbf{F_1}$ score of over \SI{80}{\%} in a challenging swapping task.
\end{abstract}

\section{Introduction}

Unmanned Aerial Vehicles (UAVs) are used in a wide range of applications including search and rescue, inspection and transportation.
Nano multirotors are small sized and lightweight UAVs (less than $\SI{50}{g}$). They are useful research tools due to their compact size, maneuverability, low cost, and swarm capabilities.
At the same time, the weight limit and computational constraints of such small platforms pose significant challenges for perception and decision making.
In the target applications, we are often interested in operating a team of nano UAVs together to increase the efficiency and robustness of a single robot.
In such a case, multiple robots have to operate in a shared space in close proximity to each other, which requires each robot to have a spatial awareness of neighboring robots to avoid dangerous formations that could lead to crashes.
Compared to ground robots, unsafe formations are difficult to detect, because they are not only caused by robots physically touching each other, but by a force created by the airflow between robots known as the \emph{downwash} effect. When one robot flies underneath another, it may suffer from a large aerodynamic force. This aerodynamic interaction may cause the crash of the lower robot. 
Thus, spatial awareness of each robot is crucial during the operation for safety reasons.

One of the most important prerequisites to operate aerial swarms safely is relative position estimation between peer robots. 
Most cooperative operations rely on external positioning systems like a motion capture system or the Global Positioning System (GPS) for this purpose. However, the former method is limited to indoor operations, while the latter fails in GPS-denied environments. Alternative solutions based on sound or infrared have been proposed \cite{audio,infrared}, but these systems require relatively large robots or specialized hardware. Infrastructure-free, accurate relative localization with weight-, power-, and budget-constraints is an important problem to be addressed. Moreover, a relative position estimator needs to be able to detect neighboring robots even when they are flying above the robot.

\begin{figure}
    \centering
    \includegraphics[width=0.8\linewidth]{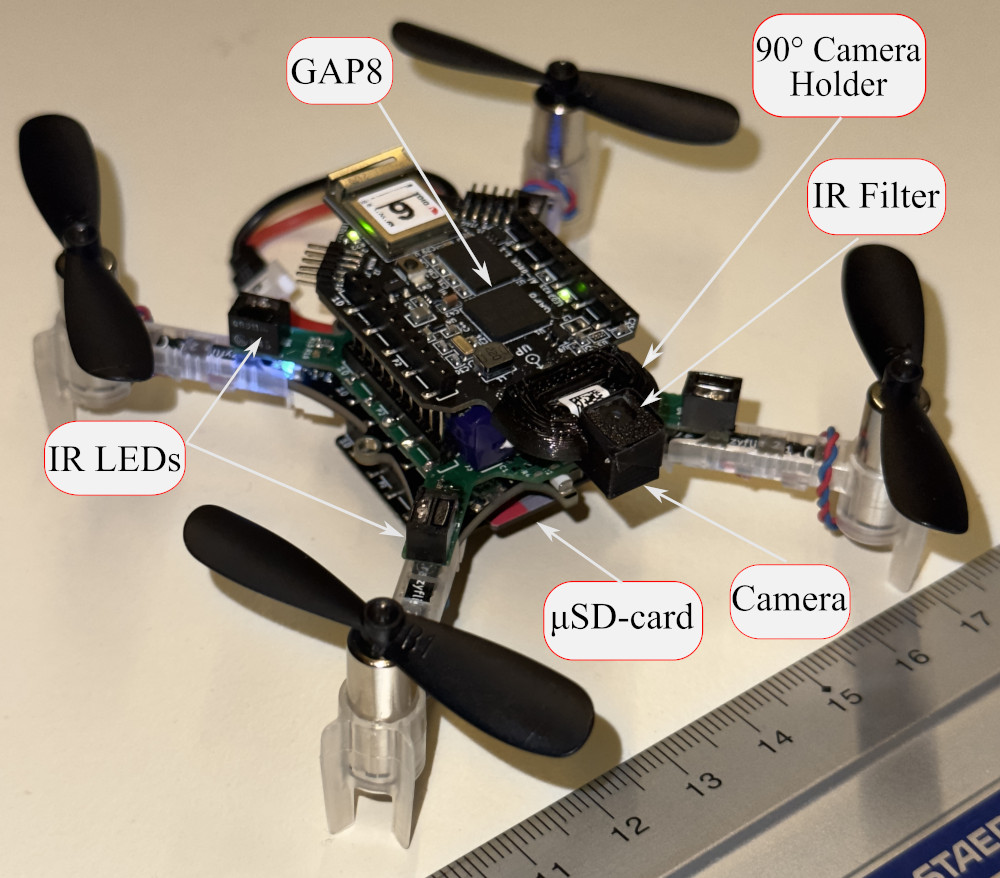}
    \caption{Our robot platform, Bitcraze Crazyflie 2.1, with custom changes: extension boards for a camera and data logging, a 3D-printed camera holder with added IR filter, and four infrared LEDs for active motion capture tracking. The total weight is \SI{42}{g} and the ruler for scale shows \si{cm}.}
    \label{fig:cf-real}
\end{figure}

In this paper, we propose to use the often neglected yaw degree of freedom to rotate a single monocular camera at rates up to \SI{8}{rad/s}, providing an affordable \emph{virtual} omnidirectional awareness instead of using multiple cameras around the robot~\cite{schilling}.
In order to observe neighboring robots above, we consider three different fixed-pitch camera angles.
Such a technique reduces the weight, cost, and computational requirements compared to prior work, while not requiring significantly more energy for the robot operation.
We focus on the use-case of predicting the presence of downwash forces directly from the obtained camera images. We use rolling shutter cameras for data collection due to its size, weight, and affordable price.

To this end, we first extend the Bitcraze Crazyflie 2.1 robotic platform to enable virtual omnidirectional perception using a low-resolution rolling-shutter camera for downwash awareness by exploiting its controllable yaw degree of freedom, see \cref{fig:cf-real}. Second, we present a new dataset with focus on close-proximity and accurate ground truth for relative position between neighbors.
The dataset consists of 3D-rendered synthetic images, as well as real flight data collected with up to four nano quadrotors in a motion capture space, one sample is given in \cref{fig:dataset-samples}.
Third, we benchmark state-of-the-art learning-based relative localization methods using our dataset and validate the results using a downwash prediction task. 
Our code and dataset are publicly available.

\section{Related Work}
\textbf{Aerodynamic interaction between UAVs.}
Neighboring UAVs in close-proximity flight experience residual forces caused by aerodynamic interactions between team members.
These forces can be measured in a lab setting using two side-by-side rotors~\cite{shukla2018} or with respect to the ground~\cite{yeo2015}, and have been shown to affect the rotor performance negatively.
One strategy of avoiding these aerodynamic disturbances is to conservatively approximate the shape of each UAV as an ellipsoid or cylinder with a large z-radii~\cite{downwash,mellinger2012}.
There are two main approaches to estimate the disturbance force: model-based~\cite{jain2019} and learning-based~\cite{shi2019, neuralswarm2, smith2023}.
The former method predicts forces using a propeller velocity field-based model, while the latter uses deep neural networks (DNNs) to capture aerodynamic interactions.
In case of known or predicted residual force, the controller of the UAV can compensate while ensuring stability.
For example, Neural Lander~\cite{shi2019} presents a learning-based controller for safe landing with account for the aerodynamic effect between a single UAV and the ground~\cite{kan2019} and NeuralSwarm2 uses a permutation invariant DNN to compensate for predicted forces within heterogeneous robot teams~\cite{neuralswarm2}.

\textbf{Multi-UAV vision-based relative localization.} 
Vision-based relative localization methods rely on onboard cameras and do not require any communication between team members.
These methods are infrastructure-free, low-cost, work with both indoor and outdoor settings, and are usable for multi-robot relative localization. 
Relative position and orientation estimation between two robots with common environmental features is proposed in \cite{montijano}.
However, this method only works when there is enough overlap between images from different cameras and if the number of features in the environment is high. Some works use black and white circular fiducial markers for the mutual localization of robots~\cite{faigl,krajnk,saska}. The main disadvantages of these methods are the need to carry printed markers and their vulnerability to lighting conditions. It is possible to use a less common wavelength to be more robust to environmental illuminations, by relying on ultraviolet (UV) LEDs~\cite{walter}. UV markers are easy to detect with UV-sensitive cameras and allow the estimation of relative position and orientation. 

Recently, techniques using convolutional neural networks (CNNs) for multi-UAV relative localization have been presented.
In these settings, CNNs detect neighboring drones from either RGB~\cite{vrba} or depth images~\cite{carrio}, then the 3D relative position of the detected neighbors can be retrieved in a post-processing step using the predicted bounding boxes.
Alternatively, learning-based methods can be trained in an end-to-end manner to predict relative positions directly from images~\cite{bonato2023}.
For our comparative study, we compare two state-of-the-art methods~\cite{vrba,locanet}, which we present in more detail in \cref{background}.

The limited field-of-view of cameras still remains a main drawback.  This shortage is addressed in~\cite{schilling}, where authors propose to mount multiple cameras on the robot in order to get omnidirectional visual input. Similarly, \cite{peihan} tackles this restricted field-of-view problem by rotating the camera with a servo motor to observe the whole environment. These approaches require aerial robots that are capable of carrying large payloads. However, UAVs that we are interested in with weight- or power constraints cannot afford heavy sensors onboard in order to have the full scene awareness.

\section{Background} \label{background}

\begin{figure}%
    \centering
    \includegraphics[width=0.49\linewidth]{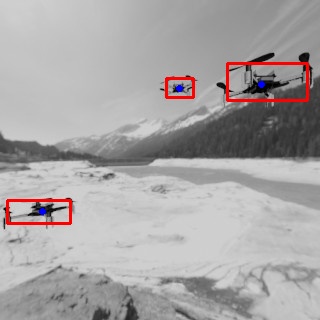}
    \hfill
    \includegraphics[width=0.49\linewidth]{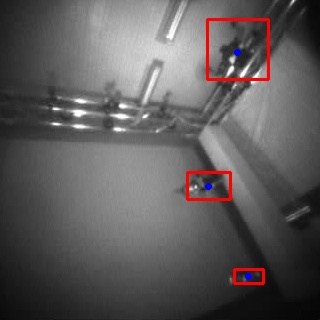}
    \caption{Examples from our dataset. Left: Synthetic image generated using Blender and Crazyswarm2. Right: Real image from flight experiments. Our dataset also contains the ground truth pose information of all robots as well as camera calibration information.}%
    \label{fig:dataset-samples}%
\end{figure}
\subsection{Single Multirotor Dynamics}
The dynamics of a single quadrotor is a six degrees-of-freedom rigid body with mass $\um$ and moment of inertia matrix $\mathbf{J}$. The quadrotor's state consists of the global position $\vp\in\mathbb R^3$, global velocity $\vv\in\mathbb R^3$, attitude rotation matrix $R\in SO(3)$ and body angular velocity $\vomega \in\mathbb R^3$. The Newton-Euler equations~\cite{mellinger2011} can be used to write the dynamics: 
\begin{subequations}
\begin{align}
    \label{eq:dynamics}
    \dot{\vp} &= \vv, & \um\dot{\vv} &= \um\vg +  R\vf_u + \vf_a,  \\
    \dot{R} &= RS(\vomega), & \mathbf{J}\dot{\vomega} &= \mathbf{J}\vomega \times \vomega + \vtau_u,
\end{align}
\end{subequations}
where $S(\cdot)$ is a skew-symmetric mapping; $\vg = (0;0;-g)^\top$ is the gravity vector; $\vf_u = (0;0;f)^\top$ and $\vtau_u = (\tau_x;\tau_y;\tau_z)^\top$ are total thrust and body torques produced by the rotors; and $\vf_a$ is the (typically unknown) residual force caused by other multirotors. 
The squared motor speeds $\vu = (n_1^2; n_2^2;...;n_k^2)$ for $k$ rotors can be controlled directly and each attached propeller creates a force and torque.
The total wrench vector $\veta$ is defined as $\veta = (f; \tau_x; \tau_y; \tau_z)^\top$ and directly proportional to the control input $\veta=B_0\vu$, where $B_0$ is the actuation matrix that contains the same scalar value $\kappa_f$ in the first row, i.e., $f = \kappa_f \sum_{i=1}^k n_i^2$.
Rotating the camera using the yaw degree of freedom requires a nonzero $\tau_z$ but does not change the total thrust $f$.
Since the energy required for the motors is proportional to $\sum_{i=1}^k n_i^2$~\cite{bitcrazePWMInvestigation}, this rotation will not increase the energy significantly.

\subsection{Downwash Effect} \label{sub:downwash}

A conservative approximation to keep $\vf_a$ small and bounded is to consider the safe robot volume as an axis-aligned ellipsoid~\cite{downwash}. The safety constraint between two robots placed at $\vp_{i}$ and $\vp_{j}$ in $\fW$ is then given by 
\begin{align} 
\label{eq:ellipsoid}
\begin{split}
\lVert\  \textbf{E}^{-1} ( \vp_{i}^\fW - \vp_{j}^\fW)  \rVert_2 \ge 2,
\end{split}
\end{align}
where $\mathbf{E}$  = $\diag(r_x,r_y,r_z)$ and $\mathbf{r}$ are the axis-aligned ellipsoid radii, typically where $0 < r_x = r_y \ll r_z $.
For localization methods, we expect that we can detect neighboring robots especially when inside or close to the border of this safety constraint.

\subsection{Camera Model}
We denote the camera coordinate frame with $\fC$, the robot frame with $\fR$, the world frame with $\fW$, the image frame with $\fI$, the position vector with $\vp=(x,y,z)$, and the matrix for transformation from frame $i$ to frame $j$ with $\mT_{i}^j$. 

Methods considered in this work estimate their neighbor robot's positions from images directly in the $\fC$ frame. The transformation of robot $i$ (frame $\fR_i$) to the camera frame of another robot $j$ (frame $\fR_j$) is
\begin{align}
\label{eq:transformation}
\begin{split}
\mT_{\fR_i}^{\fC_j} = \mT_{\fR_j}^{\fC_j}(\mT_{\fR_j}^\fW)^{-1}\mT_{\fR_i}^\fW,
\end{split}
\end{align} 
where $\mT_{\fR_j}^\fW$, $\mT_{\fR_j}^{\fC_j}$ are the transformations of robot $j$ to the world and to its own camera frames, respectively. The latter can be obtained with camera calibration. 
Points represented in $\fC$ can be transformed into $\fI$ using the intrinsic matrix $\mK$ as
\begin{align} 
\label{eq:intrinsics}
\begin{split}
\vp^{\fI}  = \hat h(\mK \vp^{\fC}) = \hat h \left(\begin{bmatrix}
f_x & 0 & c_x\\
0 & f_y & c_y \\
0 & 0 & 1
\end{bmatrix} \vp^{\fC}\right),
\end{split}
\end{align}
where $c_x$, $c_y$ and $f_x$, $f_y$ are principal points and focal lengths of the camera, respectively.
Here, we use $h(\cdot)$ to transform a point into homogeneous coordinates and $\hat h(\cdot)$ for the inverse operation.
The parameters of $\mK$ can be estimated numerically along with a nonlinear model to compensate for radial and other distortions by calibrating the camera with a known checkerboard pattern.

\subsection{Yolo-based Relative Localization}
We summarize Yolo-based localization, an approach that uses convolutional neural network (CNN) to detect nearby UAVs in images~\cite{vrba}. The CNN receives an RGB image as input and returns for each detected neighboring robot its bounding box in the image frame as well as a confidence score, indicating the probability that this is indeed a neighboring robot. The method utilizes tiny-YOLO \cite{yolo} for UAV detection that was trained on a manually labeled, single-class custom dataset. The relative position from the UAV detection is computed in two steps. First, the distance between the neighbor robot and the camera center is estimated using the predicted bounding boxes around the robot along with the robot's known physical dimensions. 
The distance from the camera origin to the center of the detected UAV is computed as
\begin{equation} \label{eq:vrba:dist}
d= r\csc\left({\frac{\alpha}{2}}\right), 
\end{equation}
assuming that the detected robot has a spherical shape with radius $r$.
Here, $\alpha$ is the angle between the two directional vectors $\mathbf{a_1}$ and $\mathbf{a_2}$ coming from the camera origin and intersecting the vertical edges of the bounding box, and $\csc$ is the cosecant trigonometric function. 

The relative position of a detected UAV with respect to the camera is calculated as
\begin{align} 
\label{eq:vrba:p}
\begin{split}
\vp_{\fR}^\fC = d\frac{\mathbf{a_c}}{\lVert\ \mathbf{a_c} \rVert} \text{ where } \mathbf{a_c} = \frac{\mathbf{a_1}+\mathbf{a_2}}{2}.
\end{split}
\end{align}

The accuracy of the estimated distance highly depends on the precision of the predicted bounding box size and thus is particularly poor for neighbors that are far away. 

\subsection{Locanet}
We summarize Locanet, a framework that estimates the 3D positions of neighboring robots using a deep neural network~\cite{locanet}. The proposed network architecture is largely similar to the object detector YOLOv3~\cite{redmon} with changes in the output.
In particular, rather than predicting bounding boxes for each robot, the network outputs for each robot the pixel coordinates of the robot's center and its distance to the camera directly. In order to support multi-robot detection, the neural network outputs $28\times40$ grid cells with two channels each.
The first channel is the confidence map. Grid cell locations refer to the center of the robot in the image plane, where a confidence score above a threshold indicates that a robot was detected. A score close to zero signals that no robot was detected in this part of the image. The second channel is the depth map containing predicted distances between the neighboring robot and the camera. This channel is only valid for high confidence scores.
Additional calculations are needed to estimate the 3D relative position between the detected neighboring robots and the camera. First, locations of grid cells with values larger than a threshold are extracted from the confidence map as well as predicted distances from the corresponding depth map grid cells. Then, the relative position of a neighboring robot with respect to the camera is calculated by inverting \eqref{eq:intrinsics}:
\begin{align} 
\label{eq:shushai2}\
\begin{split}
\vp_{\fR}^\fC =
\begin{bmatrix} z^\fC\left({\frac{\vp_{x}^\fI-c_x}{f_x}}\right),& z^\fC \left({\frac{\vp_{y}^\fI-c_y}{f_y}}\right),& z^\fC \end{bmatrix}.
\end{split}
\end{align}

\section{Dataset Collection}
In this section, we provide an overview of real-world and synthetic data collection used for comparative study. 

\subsection{Hardware} \label{robot}
We use Bitcraze Crazyflie 2.1 which are off-the-shelf commercially available, small and lightweight quadrotors. Each robot is additionally equipped with the following extensions: i) Aideck, an extension board consisting of a gray-scale camera (resolution: $324 \times 324$), an ESP32 for streaming images to the PC over Wi-Fi, and a GAP8 application processor; ii) a $\mu$SD-card extension to log on-board state information at a high rate; iii) four infrared LEDs (IR LEDs) for active optical tracking with a motion capture system; and iv) upgraded motors (betafpv 19000KV) to support the additional payload even in close-proximity flight when downwash occurs.
For the Aideck extension board, we use a custom 3D-printed camera holder that allows us to tilt the camera upwards and holds an infrared shortpass filter (\SI{710}{nm} cut-off wavelength, OD2, custom-cut glass). 
One of our robots is shown in \cref{fig:cf-real} and weighs \SI{42}{g}.

We originally used regular passive spherical reflective markers to track our robots and discovered two challenges.
First, when having other drones captured in the camera pictures, the markers are visible as bright white circles, making the collected data useless for learning-based approaches, since a neural network would simply learn how to detect white circles.
Second, the images had stripes that we presume are caused by the camera's rolling shutter in combination with the flashing infrared light sent by our optical motion capture systems.
Notably, the rapid flashing of the motion capture cameras cannot be disabled in the commonly used Optitrack motion capture system.
Unfortunately, even our IR filter was unable to avoid stripes in the captured images.
These challenges explain why there seems to be no large datasets for the Aideck camera with precise 2D/3D ground truth annotation.
Our hardware setup with IR LEDs on the UAVs allows us to disable the flashing IR LEDs of the motion capture cameras entirely. The added IR filters for the Aideck camera ensures that neighboring robots are not highlighted by bright white circles in the captured images.

\subsection{Dataset Definition}

Our dataset consists of the following:
i) gray-scale images ($320 \times 320$), ii) camera calibration $\mK$, iii) annotations for each image: a) robot poses in world frame, b) relative position of all visible neighbors with respect to the camera, c) pixels of their robot center, and d) bounding boxes in the image for each visible robot.
Examples are shown in \cref{fig:dataset-samples}.

 \subsection{Real-world Dataset} \label{subsection:real-world dataset}
Real-world images are obtained inside a $3.5 \times 3.5 \times 2.75$ \si{m^3} room equipped with a motion capture system consisting of eight Optitrack cameras.
On the host side, we use Crazyswarm2\footnote{\url{https://imrclab.github.io/crazyswarm2}}, which is based on Crazyswarm~\cite{crazyswarm} but uses ROS 2 to control and send commands for multiple Crazyflies. 
Since we are interested in close-proximity flight and the downwash effect between neighboring robots, data collection flights include scenes when one drone passes over another. Typical forward-facing cameras might fail to capture robots flying above.
In order to overcome this limitation, we equip some multirotors with \SI{45}{\degree} and \SI{90}{\degree}-tilted camera holders. 
For data collection we consider three flight scenarios with up to four robots. In the first one,
robots fly to random waypoints with different velocities with an angular yaw rate up to \SI{8}{rad/s}. This \textit{auto-yaw} motion allows localizing neighboring robots behind or outside the camera's field-of-view. In this case, we use the buffered Voronoi-cell-based collision avoidance that is part of the Crazyflie firmware.
For the second scenario, up to three robots hover in the air at different heights while the forth drone is teleoperated to fly around other hovering neighbors while recording images. These flights allow us to capture more images with multiple drones visible per frame.
For the third scenario, we let two multirotors starting at different heights swap their $x$ and $y$ positions, creating a direct downwash effect while the lower robot records images.
Examples are included in the supplemental video.

\begin{figure}
    \centering
    \includegraphics[width=\linewidth]{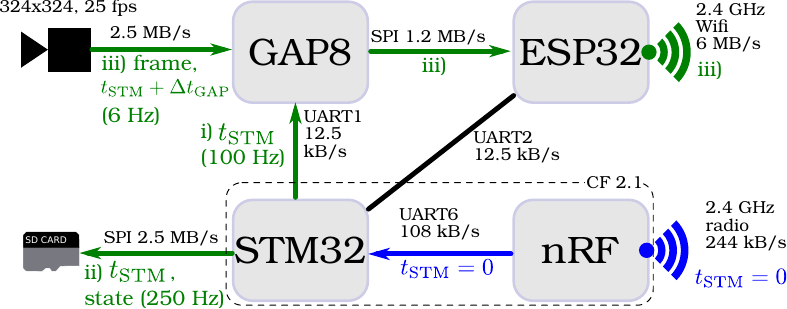}
    \caption{Microcontrollers and their connections for the robotic platform we use. The main flight controller runs on the STM, communication with a PC is done via the nRF, the camera is connected to the GAP8, which connects to the ESP to stream images over Wifi. Green indicates our process for state annotation and blue the communication for time synchronization.}
    \label{fig:architecture}
\end{figure}

We needed significant firmware changes to annotate frames with their ground truth state.
As shown in \cref{fig:architecture}, each UAV consists of four major microcontrollers with different tasks that are interconnected with each other, often with low-bandwidth interfaces such as UART.
For example, the camera is connected to the GAP8 chip, but the images are sent over WiFi from another chip (ESP32).
The speed of the Wi-Fi and the interconnect between GAP8 and ESP32 do not allow streaming images at the camera's nominal frame rate. Also, we use manual camera settings for exposure and gains. We found that using an analog gain of eight in combination with a low exposure mostly avoids motion blur in our flight scenarios. Due to the limited connection bandwidth, we stream images at about \SI{6}{fps}.
We made the following firmware changes.

\subsubsection{State Annotation} The motion capture data is sent to each robot using a custom radio, arriving at the nRF chip. However, the robot's current state estimate is needed whenever an image frame is captured at the GAP8 chip. In order to obtain correct states for each captured image, we i) send the current timestamp from the STM to GAP8 at \SI{100}{Hz}, ii) record the current state estimate and motion capture data on-board a $\mu$SD-card connected to the STM, iii) record its capture timestamp (STM clock + GAP8 since last STM clock message), and iv) use state interpolation in post-processing to compute an accurate estimate of the state.
Part i) was needed since the GAP8 has a non-negligible clock drift, even considering our short flight times.

\subsubsection{Time Synchronization} The time between different robots was synchronized by using an initial broadcast message to reset the clock, similar to~\cite{neuralswarm2}. In order to verify the synchronization between two STM32's, we rigidly mount both robots at the end of a teeterboard and teeter the board repeatedly.
For each robot, the gyroscope data was recorded with the associated STM32 timestamp to a $\mu$SD-card.
The teeter events show up as a sharp gyroscope spike and we compute the optimal temporal offset that minimizes the error between the gyroscope sequences in four cases. The obtained time delay between two STM32's is \SI{0.6}{ms} ($\pm$ \SI{0.2}{ms}).
Time synchronization between the camera and the STM32 clock is verified with a similar experiment.
The timestamp for the start of the gyroscope spikes are aligned with image timestamps capturing the robot motion change.
Over five trials, we obtained \SI{16.8}{ms} ($\pm$ \SI{20}{ms}).
Our camera records at \SI{30}{fps} and can capture an image every \SI{33}{ms}.
The expected time delay as computed with our method is \SI{16.5}{ms} due to the low camera sampling rate. Thus, the computed clock difference is negligible.

\subsection{Synthetic Dataset}
We extend the software-in-the-loop simulation of Crazyswarm2 with a plugin to support the 3D rendering software Blender.
This allows us to use the same high-level flight scripts with random waypoints as the real drones.
Each robot can be equipped with virtual cameras with different mounting angle and perform auto-yaw rotation while recording images in our simulated environment.
In order to minimize the gap between synthetic and real-world images, we match the robots' physical dimensions and camera settings.
We use publicly available panorama pictures for background augmentation.

\subsection{Automatic Dataset Annotation}
We contribute an automated approach to annotate each image with relative position in $\mathcal{C}$, as well as robot center locations and bounding boxes in $\mathcal{I}$ for each visible robot. Our auto-labeling process pipeline starts with linear interpolation of the robot's position and orientation using the camera's timestamp for each taken frame. Pose interpolation ensures that the position and orientation of the robot depicted in the images are close to the actual pose when the frame was captured. Afterwards, we retrieve each neighboring robot's position with respect to the camera with \eqref{eq:transformation}.  
The robot center pixels are estimated with \eqref{eq:intrinsics} and those outside the field-of-view are filtered. 

We assume that the physical dimensions of the neighboring robot are known apriori, and construct eight points $\vrho_i (i = 1,\ldots,8)$ representing the corners of the robot in $\fR$.
We transform these corners into the camera frame and project to the image using \eqref{eq:intrinsics}.
Then, the bounding box corners are given by the resulting maximum and minimum values.
Formally,
\begin{align}
\label{eq:bb}
\begin{split}
\vb_{\text{min}}, \vb_{\text{max}} = \minmax_{\vrho_i} \hat h(\mK \hat h(\mT_{\fR}^{\fC} h(\vrho_i))).
\end{split}
\end{align}

We use three major steps to ensure the accuracy of $\mK$ and $\mT_{\fR}^{\fC}$. First, the camera intrinsic parameters are estimated with the OpenCV library using a checkerboard and only first-order distortion estimation due to the low resolution of our cameras. Second, for each collected dataset, we use an exhaustive enumeration of the discretized camera rotation vector to compute the correct camera-to-robot orientation. This second step is necessary for precise annotations, because our camera is only semi-rigidly attached to the robot frame: the Aideck is only pinned on top and the camera may shift slightly in its 3D-printed frame.
Finally, we manually inspect real-world images and filter outlier frames. 
  
\section{Benchmark Results}
For our comparative study, we focus on two learning-based multi-robot relative localization frameworks from \cref{background}.

\subsection{Evaluation}
\subsubsection{Synthetic Images}
\begin{figure}%
    \centering
    \includegraphics[height=4.2cm,page=1]{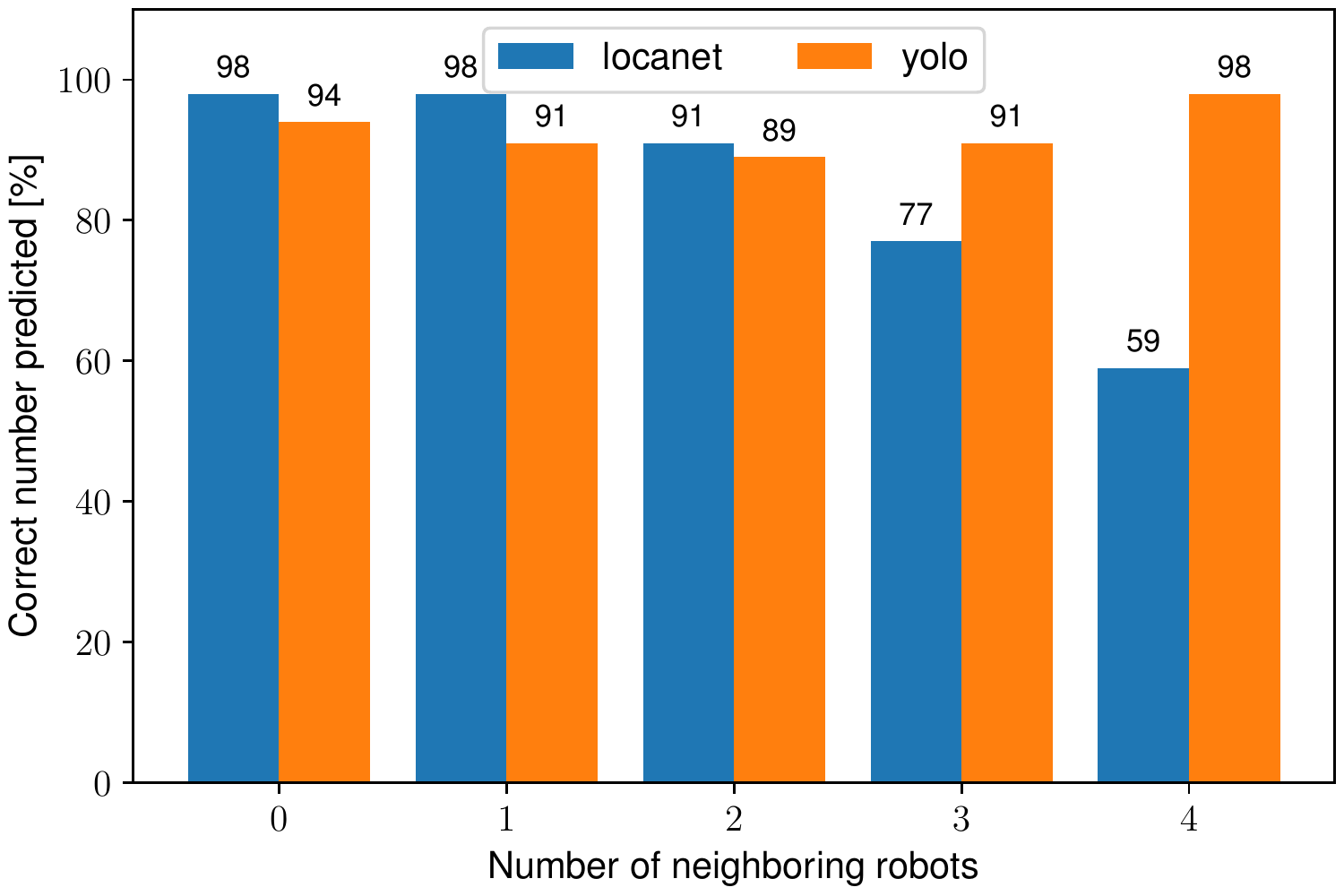}\includegraphics[height=4.2cm,page=3]{figs/result_syn.pdf}
    \caption{Results for synthetic images. Left: success rate. Right: Whisker plot for the Euclidean error for all cases where both methods predicted the number of neighbors correctly.}%
    \label{fig:results-synth}%
\end{figure}

Both neural network models are trained from scratch on 50000 synthetic images for 50 epochs. The testing dataset consists of 500 synthetic images (unseen; same domain as training). The number of images are equally distributed for zero, one, two, three, and four visible neighboring robots.
\cref{fig:results-synth} shows the prediction precision and accuracy of the estimated positions.
For the precision, we report the percentage of images where the neural networks correctly predicted the number of visible neighbors.
For the accuracy, we show the distribution of Euclidean errors between ground truth relative position and predicted robot position.
For evaluation, we only include images where both models correctly predicted the number of visible robots.
For the images with multiple robots, we first find the optimal assignment of prediction and ground truth robot position using the Hungarian method, and then include the resulting individual assignment costs in the reported distribution.

\begin{figure}%
    \centering
    \includegraphics[height=4.2cm,page=1]{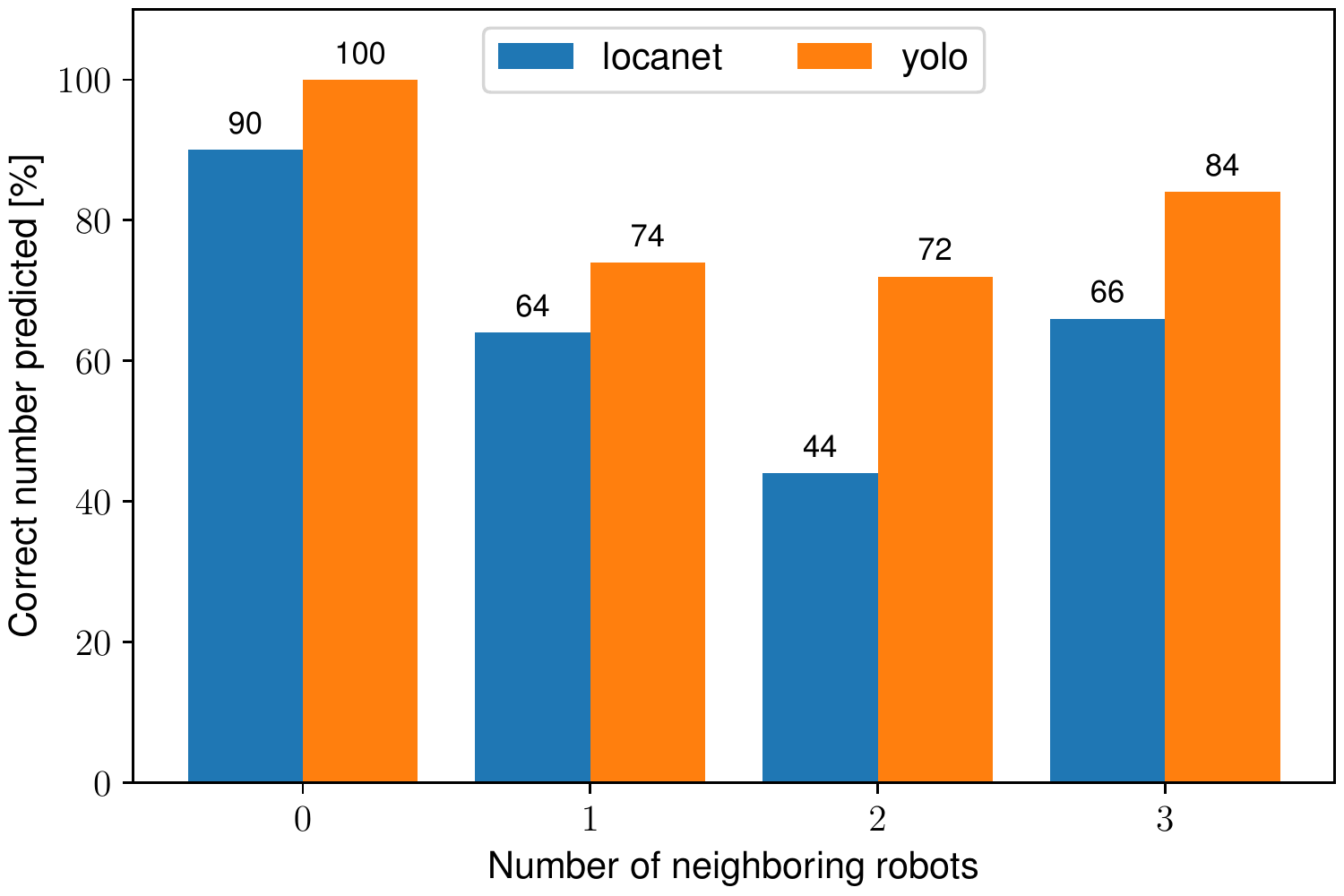}\includegraphics[height=4.2cm,page=3]{figs/result_real.pdf}
    \caption{Results for real images. Left: success rate. Right: Whisker plot for the Euclidean error for all cases where both methods predicted the number of neighbors correctly.}%
    \label{fig:results-real}%
\end{figure}

\begin{table*}

\caption{Downwash prediction on real images ($F_1$-score; precision, recall in gray). Bold entries are best for the row (ground truth) or columns (Locanet and Yolo).}
\centering
\setlength{\tabcolsep}{0.3em}
\begin{tabular}{l||l|l|l||l|l|l||l|l|l||l|l|l}
Auto-Yaw & \multicolumn{3}{c||}{\SI{2}{rad/s}} & \multicolumn{3}{c||}{\SI{4}{rad/s}} & \multicolumn{3}{c||}{\SI{6}{rad/s}} & \multicolumn{3}{c}{\SI{8}{rad/s}}\\
\hline
Camera  & Forward & $45^{\circ}$ & Up & Forward & $45^{\circ}$ & Up & Forward & $45^{\circ}$ & Up & Forward & $45^{\circ}$ & Up\\
\hline\hline
ground truth & 0.14\,{\color{gray}\tiny 0.1}\,{\color{gray}\tiny 0.4}  & 0.45\,{\color{gray}\tiny 0.6}\,{\color{gray}\tiny 0.4}  & 0.99\,{\color{gray}\tiny 1.0}\,{\color{gray}\tiny 1.0}  & 0.10\,{\color{gray}\tiny 0.1}\,{\color{gray}\tiny 0.2}  & 0.52\,{\color{gray}\tiny 0.6}\,{\color{gray}\tiny 0.5}  & \textbf{1.00\,{\color{gray}\tiny 1.0}\,{\color{gray}\tiny 1.0} } & 0.38\,{\color{gray}\tiny 0.3}\,{\color{gray}\tiny 0.6}  & 0.47\,{\color{gray}\tiny 0.7}\,{\color{gray}\tiny 0.4}  & \textbf{1.00\,{\color{gray}\tiny 1.0}\,{\color{gray}\tiny 1.0} } & 0.30\,{\color{gray}\tiny 0.2}\,{\color{gray}\tiny 0.5}  & 0.56\,{\color{gray}\tiny 0.8}\,{\color{gray}\tiny 0.4}  & \textbf{1.00\,{\color{gray}\tiny 1.0}\,{\color{gray}\tiny 1.0} }\\
\hline\hline
Locanet & \textbf{0.16\,{\color{gray}\tiny 0.1}\,{\color{gray}\tiny 0.2} } & 0.00\,{\color{gray}\tiny nan}\,{\color{gray}\tiny 0.0}  & 0.48\,{\color{gray}\tiny 1.0}\,{\color{gray}\tiny 0.3}  & 0.00\,{\color{gray}\tiny nan}\,{\color{gray}\tiny 0.0}  & 0.05\,{\color{gray}\tiny 1.0}\,{\color{gray}\tiny 0.0}  & 0.59\,{\color{gray}\tiny 0.9}\,{\color{gray}\tiny 0.4}  & \textbf{0.16\,{\color{gray}\tiny 0.2}\,{\color{gray}\tiny 0.2} } & 0.00\,{\color{gray}\tiny nan}\,{\color{gray}\tiny 0.0}  & 0.86\,{\color{gray}\tiny 1.0}\,{\color{gray}\tiny 0.8}  & \textbf{0.22\,{\color{gray}\tiny 0.2}\,{\color{gray}\tiny 0.3} } & 0.05\,{\color{gray}\tiny 0.2}\,{\color{gray}\tiny 0.0}  & 0.50\,{\color{gray}\tiny 1.0}\,{\color{gray}\tiny 0.3} \\
Yolo & 0.06\,{\color{gray}\tiny 0.0}\,{\color{gray}\tiny 0.1}  & \textbf{0.47\,{\color{gray}\tiny 0.7}\,{\color{gray}\tiny 0.4} } & \textbf{0.83\,{\color{gray}\tiny 1.0}\,{\color{gray}\tiny 0.7} } & \textbf{0.14\,{\color{gray}\tiny 0.1}\,{\color{gray}\tiny 0.2} } & \textbf{0.51\,{\color{gray}\tiny 0.6}\,{\color{gray}\tiny 0.5} } & \textbf{0.82\,{\color{gray}\tiny 1.0}\,{\color{gray}\tiny 0.7} } & 0.08\,{\color{gray}\tiny 0.1}\,{\color{gray}\tiny 0.1}  & \textbf{0.49\,{\color{gray}\tiny 0.8}\,{\color{gray}\tiny 0.3} } & \textbf{0.92\,{\color{gray}\tiny 1.0}\,{\color{gray}\tiny 0.9} } & 0.21\,{\color{gray}\tiny 0.2}\,{\color{gray}\tiny 0.2}  & \textbf{0.57\,{\color{gray}\tiny 0.8}\,{\color{gray}\tiny 0.5} } & \textbf{0.83\,{\color{gray}\tiny 1.0}\,{\color{gray}\tiny 0.7} }\\
\end{tabular}

\label{tab:downwash}
\end{table*}

\subsubsection{Real Images}
We train both models with 8216 real images in total for 50 epochs, starting from the weights from the synthetic dataset training.
Images acquired during the first two flight scenarios from \cref{subsection:real-world dataset} are used for the training, and its subset of 250 unseen images are used to evaluate the success rate and relative position estimation. 
The success rate and Euclidean error distributions are computed similar to evaluation with synthetic images and given in \cref{fig:results-real}.

\subsubsection{Downwash Awareness}
We analyze the ability of our rotating, fixed-pitch camera to correctly predict downwash with a dataset, where two drones at different heights swap their $x$ and $y$ positions. Angular yaw rate and the camera angles are changed for each flight, see \cref{tab:downwash}.

To predict the downwash for each image, we use the following steps.
First, for each image we compute a \emph{belief set} $\sB \in \mathbb{P}(\mathbb R^3)$, where $\mathbb P$ is the powerset, containing the positions of neighboring robots in $\fW$, by i) sorting images by their timestamp, ii) setting the initial belief to $\sB = \emptyset$, iii) for each image we try to project $\vp_i\in\sB$ to the image and remove $\vp_i$ from $\sB$ if the reprojection is successful, and we add all visible robots as predicted to $\sB$.
Second, for each image we compute if downwash occurred by evaluating condition \eqref{eq:ellipsoid} using $\vp_i^\fW$ and $\vp_j^\fW$ from the ground truth data and $\mathbf{E} = \diag(0.15, 0.15, 0.3)$ \si{m}.
Third, we compute if downwash is predicted using $\sB$ and condition \eqref{eq:ellipsoid} using the predicted positions transformed to $\fW$.
Finally, we report the $F_1$-score, precision, and recall for different cases.

\subsection{Discussion}
For the synthetic images, the percentage of predictions with the correct number of robots for Locanet is slightly higher than the Yolo-based approach for images with up to two visible drones as shown in \cref{fig:results-synth}. However, for cases with three or more visible robots, the success rate of Locanet drops significantly, while the Yolo-based method shows a slight improvement. This might be due to cases where two robots occupy the same neural network output grid cell, and the confidence map of Locanet returns depth prediction for just one of the robots while ignoring another. 

For real images the success rate of both models to predict the correct number of visible robots decreases compared to synthetic images. Overall, the Yolo-based method demonstrates higher and more consistent success rates.

The distribution of Euclidean errors between ground truth and predicted relative positions for synthetic images shows that the mean error of Locanet is lower than of Yolo. Moreover, the outliers are worse for the Yolo-based method. When the width of the predicted bounding boxes are less accurate, mistakes of only a few pixels can lead to largely different distance estimations (e.g., $\pm$2 pixel error causes Euclidean error of \SI{0.2}{m}). Surprisingly, for real-world images, we observe the opposite effect, i.e., Yolo has a lower mean and fewer outliers compared to Locanet.
However, this result is not statistically significant due to the high standard deviation of both methods.

Downwash awareness of the two considered models on real images is summarized in \cref{tab:downwash}. Here, we compare 12 possible combinations (4 angular yaw velocities and 3 camera tilt angles). Downwash prediction while rotating the camera with different angular rates works best with the upward-facing camera for both methods achieving an $F_1$ score of over \SI{80}{\percent} with the Yolo-based approach. Both models achieve the highest $F_1$ score over all test cases when the camera is rotating with \SI{6}{rad/s}.
The $45^{\circ}$ tilted camera is not predicting the downwash accurately even when using ground truth prediction ($F_1$ score of less than \SI{60}{\percent}).
Nevertheless, the Yolo-based approach is able to match this theoretical upper bound.
A simple forward-facing camera is unusable for the downwash prediction task ($F_1$ score of less than \SI{30}{\percent} for ground truth prediction and real predictions).

\section{Conclusion}

We present a novel way to provide multirotors with omnidirectional vision by using the yaw degree of freedom to rotate a fixedly mounted camera. 
In this setting, we create a new dataset of 50k synthetic and 9k real images, annotated with accurate ground truth poses in world frame, positions in camera frame, and bounding boxes in image frame. Based on our dataset, we benchmark two state-of-the-art learning-based relative localization methods with respect to success rate, relative position estimation, and downwash prediction accuracy. Our results show that an upward-facing rotating camera detects all flying neighbors when assuming perfect perception and that Yolo-based relative localization still achieves an $F_1$ score of over \SI{80}{\%}, even at fairly low angular rates of \SI{2}{rad/s}.
In future research, we will demonstrate that our omnidirectional vision can be used for on-board autonomy such as collision avoidance between robots.

\bibliographystyle{IEEEtran}
\bibliography{IEEEabrv,references}

\end{document}